\documentclass{article}

\usepackage[preprint]{neurips_2025}

\usepackage[utf8]{inputenc} % allow utf-8 input
\usepackage[T1]{fontenc}    % use 8-bit T1 fonts
\usepackage{hyperref}       % hyperlinks
\usepackage{url}            % simple URL typesetting
\usepackage{booktabs}       % professional-quality tables
\usepackage{amsfonts}       % blackboard math symbols
\usepackage{nicefrac}       % compact symbols for 1/2, etc.
\usepackage{microtype}      % microtypography
\usepackage{xcolor}         % colors
\usepackage{graphicx}
\usepackage{algorithm}
\usepackage{algorithmic}

\usepackage{amsmath}
\usepackage{amssymb}
\usepackage{booktabs}
\usepackage{pifont}
\usepackage{times}
\usepackage{multirow}
\usepackage[utf8]{inputenc}
\usepackage{xcolor}
\usepackage{pifont}
\usepackage{multirow}  % For multirow cells
\usepackage{pifont}    % For checkmarks and x marks
\usepackage{newfloat}
\usepackage{listings}
\usepackage{tcolorbox}
\usepackage{fontawesome5}
\usepackage{ulem}

\title{GroundedPRM: Tree-Guided and Fidelity-Aware Process Reward Modeling for Step-Level Reasoning}

\author{Yao Zhang \textsuperscript{\rm 1,6} \thanks{Corresponding authors: yzhang@dbs.ifi.lmu.de, hanzhen02111@163.com}
\hspace{0.2em}  \thanks{Equal contribution}
\qquad Yu Wu \textsuperscript{\rm 2} 
\footnotemark[2]
\qquad Haowei Zhang \textsuperscript{\rm 3} \thanks{Work done during an internship at Technical University of Munich}
\qquad Weiguo Li \textsuperscript{\rm 4}
\qquad Haokun Chen\textsuperscript{\rm 1} \\
\qquad \bf Jingpei Wu\textsuperscript{\rm 1} 
\qquad \bf Guohao Li\textsuperscript{\rm 5} 
\qquad \bf Zhen Han \textsuperscript{\rm 1} \footnotemark[1] \qquad \bf Volker Tresp\textsuperscript{\rm 1,6}\\
    \textsuperscript{\rm 1} LMU Munich \hspace{0.2em}
    \textsuperscript{\rm 2} Technical University of Munich \hspace{0.2em}
    \textsuperscript{\rm 3} Fudan University \hspace{0.2em}
    \textsuperscript{\rm 4} University Heidelberg \\
    \textsuperscript{\rm 5} University of Oxford \qquad
    \textsuperscript{\rm 6} Munich Center for Machine Learning \\
   	}

\begin{document}

\maketitle

\begin{abstract}
Process Reward Models (PRMs) aim to improve multi-step reasoning in Large Language Models (LLMs) by supervising intermediate steps and identifying errors throughout the reasoning process. However, building effective PRMs remains challenging due to the lack of scalable, high-quality annotations. Existing approaches rely on costly human labeling, LLM-based self-evaluation that is prone to hallucination, or Monte Carlo (MC) estimation, which infers step quality solely from rollout outcomes and often introduces noisy, misaligned supervision due to credit misattribution. These issues result in three core limitations: noisy rewards, low factual fidelity, and misalignment with step-level reasoning objectives. To address these challenges, we introduce \textbf{GroundedPRM}, a tree-guided and fidelity-aware framework for automatic process supervision. To reduce reward noise and enable fine-grained credit assignment, we construct structured reasoning paths via Monte Carlo Tree Search (MCTS). To eliminate hallucinated supervision, we validate each intermediate step using an external tool, providing precise, execution-grounded correctness signals. To combine both step-level validation and global outcome assessment, we design a hybrid reward aggregation mechanism that fuses tool-based verification with MCTS-derived feedback. Finally, we format the reward signal into a rationale-enhanced, generative structure to promote interpretability and compatibility with instruction-tuned LLMs. GroundedPRM is trained on only 40K automatically labeled samples, amounting to just \textbf{10\%} of the data used by the best-performing PRM trained with auto-labeled supervision. Nevertheless, it achieves up to a \textbf{26\% relative improvement} in average performance on ProcessBench. When used for reward-guided greedy search, GroundedPRM outperforms even PRMs trained with human-labeled supervision, offering a scalable and verifiable path toward high-quality process-level reasoning. Our code is publicly released at \href{https://yaoz720.github.io/GroundedPRM/}{github.com/GroundedPRM}.

\end{abstract}

\section{Introduction}
Large Language Models (LLMs)~\cite{achiam2023gpt, guo2025deepseek, team2023gemini} have demonstrated impressive capabilities in planning~\cite{huang2024understandingplanningllmagents, zhang2025swarmagentic}, decision-making~\cite{li2024embodied}, and complex task execution~\cite{xi2024agentgym, zhang2025webpilot}. However, they remain prone to hallucinations and reasoning errors, particularly in multi-step tasks such as mathematical problem solving. Existing methods like Chain-of-Thought prompting~\cite{wei2022chain, yang2024qwen2} and Test-Time Scaling~\cite{liu2025can, snell2025scaling} improve final accuracy, yet LLMs often produce solutions that appear coherent while containing errors in reasoning or calculation. These issues are further exacerbated by outcome-level supervision and coarse decoding strategies, e.g., majority voting, which overlook step-level correctness and provide little guidance during intermediate reasoning.

To mitigate these shortcomings, Process Reward Models (PRMs) have emerged as a promising direction~\cite{lightman2023let}. PRMs assign step-level scores to reasoning trajectories, enabling fine-grained supervision that supports better control and interpretability in multi-step reasoning. However, developing effective PRMs remains challenging due to the lack of reliable and faithful reward signals for training. Human annotation~\cite{lightman2023let}, while accurate, is costly and unscalable. LLM-as-a-judge~\cite{zheng2023judging} is more efficient but susceptible to hallucination, often rewarding fluent yet incorrect reasoning and thus compromising factual fidelity. Monte Carlo (MC) estimation~\cite{luo2024improve, wang2023math} provides another alternative by inferring step quality from final rollout outcomes, but it introduces noisy reward due to credit misattribution: correct steps may be penalized if the rollout fails, while flawed steps may be rewarded if the final answer happens to be correct~\cite{zhang2025lessons}. Moreover, MC estimation typically evaluates only final outcomes, ignoring explicit assessment of intermediate step correctness, which misaligns the supervision signal with the objective of step-wise reasoning accuracy.

Several recent works have attempted to refine MC-based supervision, but core limitations persist. OmegaPRM~\cite{luo2024improve} uses a binary search strategy to locate the first incorrect step, but still relies on rollout success to infer correctness, leaving credit assignment coarse. Qwen2.5-Math-PRM~\cite{zhang2025lessons} filters samples based on agreement between MC estimation and LLM judgments, but this strategy inherits hallucination bias and scores each step solely based on rollout outcomes, without assessing whether it contributes to or hinders correct reasoning. BiRM~\cite{chen2025better} augments PRM with a value head to predict future success probability, but both reward and value signals are derived from noisy rollouts and lack external validation. These approaches offer partial improvements, yet remain constrained by outcome-based heuristics, hallucination-prone feedback, or weak step-level credit modeling.

To address these challenges, we propose \textbf{GroundedPRM}, a tree-guided and fidelity-aware framework for automatic process supervision. GroundedPRM is designed to resolve three core limitations in existing PRMs: noisy rewards, low factual fidelity, and misalignment with step-level reasoning objectives. First, to reduce reward noise and improve credit attribution, GroundedPRM leverages Monte Carlo Tree Search (MCTS) to construct structured reasoning paths and assess each step based on its contribution within the trajectory. Second, to ensure factual grounding, each intermediate step is verified using an external math tool, producing correctness signals based on executable logic rather than LLM-generated feedback, thereby eliminating hallucinated supervision. Third, to combine step-level validation with global outcome assessment, we design a hybrid reward aggregation mechanism that fuses tool-based verification with MCTS-derived feedback. Finally, all rewards are formatted into binary decisions paired with rationale-enhanced justifications, enabling interpretable supervision signals that are compatible with LLM-based generation and downstream reasoning workflows.

We evaluate GroundedPRM on ProcessBench and observe substantial gains in both data efficiency and overall performance. It is trained on only 40K automatically labeled samples, just \textbf{10\%} of the data used by the best-performing PRM trained with auto-labeled supervision, yet achieves up to a \textbf{26\% relative improvement} in average performance. Furthermore, when deployed in reward-guided greedy search, where candidate steps are selected based on predicted reward, GroundedPRM surpasses even PRMs trained with human-labeled supervision, establishing new state-of-the-art results across multiple mathematical reasoning benchmarks. These findings highlight the effectiveness, scalability, and practical value of our structured and fidelity-aware supervision framework for both training and inference.

The key contributions of this work are:
\begin{enumerate}
	\item We propose GroundedPRM, a tree-guided and fidelity-aware process reward modeling framework that leverages MCTS to construct structured reasoning paths and support step-level credit assignment.
	\item We introduce a fidelity-aware verification mechanism that validates each reasoning step using an external math tool, ensuring correctness grounded in executable logic and eliminating hallucinated supervision.
	\item We design a hybrid reward aggregation mechanism that integrates tool-based step validation with feedback derived from MCTS-guided reasoning paths.
	\item We format rewards into a rationale-enhanced, generative structure to improve interpretability and enable seamless integration into inference-time decoding and downstream reasoning workflows.
	\item We demonstrate strong data efficiency and inference performance by evaluating GroundedPRM on ProcessBench and reward-guided greedy search.
\end{enumerate}

\section{Related Work}
\subsection{Mathematical Reasoning with LLMs}
Large Language Models (LLMs) have shown remarkable progress in solving math problems via Chain-of-Thought (CoT) reasoning, where step-by-step solutions often improve final answer accuracy~\cite{wei2022chain}. Building on this, recent efforts have focused on enhancing reasoning capabilities through pretraining on math-related corpora~\cite{jaech2024openai,shao2024deepseekmath,ying2024internlm}, instruction tuning with annotated derivations~\cite{li2024embodied, yang2024qwen2,ye2025limo, yu2023metamath}, and prompting strategies tailored for math tasks~\cite{brown2024large,imani2023mathprompter, kojima2022large}. Despite these improvements, LLMs remain vulnerable to intermediate reasoning errors, even when final answers are correct~\cite{zheng2024processbench}. This discrepancy undermines the reliability of generated solutions, motivating the use of external verification or inference-time selection strategies~\cite{guo2025deepseek, snell2024scaling, team2024qwq}. Such approaches typically operate at the output level, offering limited supervision for correcting internal steps. Unlike prior methods that intervene at the output level, our approach supervises the reasoning process itself via step-level reward modeling, enabling finer-grained error identification and ensuring more faithful alignment with step-level reasoning and factual correctness.

\subsection{Process Reward Models for Step-Level Supervision}
To enhance reasoning fidelity and identify intermediate errors, PRMs have emerged as a promising alternative to outcome-level supervision~\cite{lightman2023let,uesato2022solving}. PRMs evaluate the correctness of individual reasoning steps and have been shown to improve alignment and generalization across math tasks~\cite{wang2023math, zhang2025lessons}. A key challenge lies in generating reliable step-level annotations. Early methods rely on expert-labeled datasets such as PRM800K~\cite{lightman2023let}, which are expensive to scale. Recent work has explored automatic synthesis through MC estimation~\cite{luo2024improve, wang2023math}, often leveraging rollout outcomes to infer step validity. However, MC-based supervision introduces noise due to credit misattribution and dependency on the quality of the completion model~\cite{zhang2025lessons, zheng2024processbench}. To mitigate this, several methods combine MC with LLM-as-a-judge consensus filtering~\cite{zhang2025lessons} or adopt preference-based learning frameworks~\cite{chen2024step}. In contrast, our method GroundedPRM constructs PRM supervision from the ground up by integrating tree-structured search via MCTS~\cite{browne2012survey}, step-level verification with external math engines, and fused value–correctness reward modeling. This pipeline produces reward signals that are verifiable, structurally grounded, and directly aligned with step-level reasoning objectives, effectively addressing the fidelity and alignment issues that prior methods leave unresolved.

\begin{figure*}[t]
     \centering
    \includegraphics[width=\textwidth]{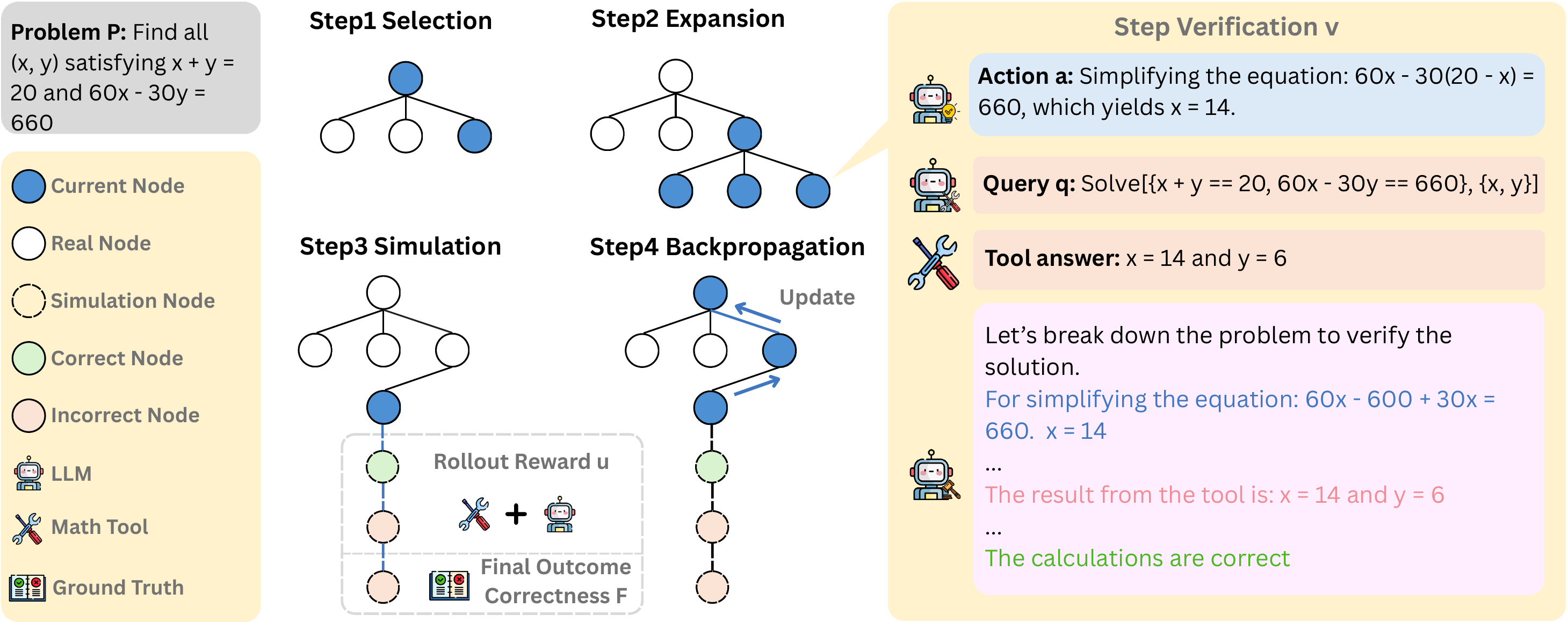}
    \caption{Overview of the GroundedPRM framework. GroundedPRM constructs reasoning paths via MCTS, where each node corresponds to an LLM-generated step. During simulation, intermediate steps are verified using an external tool, and final answers are checked against ground truth. Step-level and outcome-level correctness signals are aggregated into a rollout reward, which is backpropagated along the tree to update node statistics; the next node is then selected by UCT, continuing the MCTS search until convergence or budget exhaustion. The framework enables verifiable, interpretable, and structure-aware process supervision for multi-step reasoning. The generative rationale provides interpretable feedback for each step.}
    \label{fig:main}   
\end{figure*}

\section{Methodology}
\label{sec:method}

GroundedPRM is designed to address three core limitations of existing process reward modeling methods: noisy rewards, low factual fidelity caused by hallucinated self-assessment, and misalignment with step-level reasoning objectives. These challenges call for a framework that can assign fine-grained credit, validate the factual correctness of individual steps, and integrate local and global signals into a reliable and interpretable supervision objective. To this end, GroundedPRM introduces a tree-guided and fidelity-aware reward modeling framework composed of four core components. First, it employs Monte Carlo Tree Search (MCTS) to construct structured reasoning paths and assess each step based on its contribution within the search trajectory, enabling more stable and attribution-aware supervision than flat sampling-based methods. Second, it verifies each intermediate step using an external tool, producing binary correctness labels grounded in executable logic and thereby mitigating hallucinated feedback from the model. Third, it unifies verified step-level signals and final-answer correctness into a joint supervision objective, maintaining fine-grained credit assignment while offering stable and reasoning-grounded supervision.

Finally, the reward supervision is formatted into a rationale-enhanced generative structure, pairing each step with both a binary score and an explanation to support interpretability and compatibility with instruction-tuned LLMs. An overview of this framework is illustrated in Fig.~\ref{fig:main}. We provide the full algorithmic pseudocode in Appendix~\ref{app:code}.

\subsection{Tree-Guided Reasoning Path Construction}
\label{sec:mcts}
To enable stable and attribution-aware process supervision, GroundedPRM employs MCTS to construct structured reasoning paths for each input problem \( P \). Each node in the search tree is associated with a partial reasoning state \( s = \{s_1, \dots, s_i\} \), representing the sequence of previously generated reasoning steps. In addition to the state, each node stores auxiliary information including tool queries \( q \), verification outcomes \( v \), and value estimates \( Q \). A reasoning step is represented as an action $a$, defined as a natural language expression generated by the LLM that extends the current reasoning state, transitioning it from state $s$ to a new state $s’$. The value function \( Q(s, a) \) estimates the expected return of applying action \( a \) in state \( s \), and is updated through feedback from simulated rollouts. The search process consists of four stages:

\paragraph{Selection.} 
Starting from the root node, the algorithm recursively selects child nodes according to a tree policy until reaching a node that is not fully expanded. To balance exploration and exploitation, we use the Upper Confidence Bound for Trees (UCT)~\cite{kocsis2006bandit}, which balances estimated value with an exploration bonus that decreases as a node is visited more often, thereby encouraging the search toward promising yet under-explored nodes. The UCT score for each candidate action $a$ at state $s$ is computed as:
\begin{equation}
\mathrm{UCT}(s,a) = Q(s,a) + c \cdot \sqrt{\frac{\log N(s)}{N(s,a)}},
\end{equation}
where $N(s)$ and $N(s,a)$ are the visit counts of the parent and child nodes, respectively; and $c$ is a hyperparameter controlling the exploration strength.

\paragraph{Expansion.}
If the selected node is not terminal, it is expanded by sampling $K$ new actions from LLM, each producing a distinct child state $s'$. We set $K=3$ in our experiments. This constrains the branching factor while maintaining reasoning diversity.

\paragraph{Simulation.}
From the newly expanded node, we simulate a complete reasoning trajectory by sequentially sampling steps \( s_{i+1}, \dots, s_T \) until the model produces a final answer. We sample from the current state using the LLM in a left-to-right fashion to complete the solution. For each step \( s_j \) where $j\in\{i+1, ..., T-1\}$, we obtain a binary correctness label \( v_j \in \{-1,1\} \) using the tool-based verification procedure described in \S~\ref{sec:verification}. Additionally, the final answer is compared against the ground-truth solution to determine the overall outcome \( F \in \{-1,1\} \). We adopt signed labels \{-1,+1\} instead of \{0,1\} so that incorrect steps propagate negative feedback, thereby decreasing node values during MCTS rather than being treated as neutral. These per-step and final correctness signals are subsequently aggregated into a single rollout reward \( u_i \), as defined in \S~\ref{sec:aggregation}.

\paragraph{Backpropagation.}
The reward \( u \) computed for the simulated trajectory is propagated backward along the path traversed during selection. For each visited state-action pair \( (s_k, a_k) \) at depth \( d_k \) from the terminal node, we update its value as:
\begin{align}
Q(s_k, a_k) \leftarrow Q(s_k, a_k) + \gamma^{d_k} \cdot (u_i+v_i),
\end{align}
where $k\in\{0,...,i-1\}$, \( \gamma \in (0, 1) \) is a decay factor controlling temporal discount, and \( d_k \) denotes the number of steps from node $i$. This update scheme assigns stronger credit to steps closer to the final outcome, aligning attribution with their causal impact in the reasoning process.

By iteratively executing the four MCTS stages, GroundedPRM constructs a structured and diverse distribution over reasoning paths. This search process prioritizes trajectories with high step-level validity and globally correct outcomes, yielding supervision signals that are both structure-aware and attribution-sensitive. The resulting credit assignments are more stable and fine-grained than those produced by flat Monte Carlo rollouts, directly addressing reward noise and misattribution. Multiple rollouts are performed per input to balance path diversity with search efficiency.

\subsection{Fidelity-Aware Step Verification with External Tool}
\label{sec:verification}

To ensure reward fidelity and eliminate hallucinated supervision, GroundedPRM integrates step-level verification into each reasoning step via external tools. During simulation (see \S~\ref{sec:mcts}), the LLM generates a sequence of reasoning steps $\{s_{i+1}, \dots, s_T\}$, where each $s_j$ ($i+1 \leq j \leq T-1$) denotes an intermediate reasoning step expressed in natural language during rollout.

For each step \( s_j \), we construct a corresponding structured math query and submit it to an external math tool, such as Wolfram Alpha (WA)~\cite{wolframalpha2025}. The tool's response is parsed to determine whether the computation or transformation expressed in \( s_j \) is factually correct. We represent this outcome as a binary verification label \( v_j \in \{-1,1\} \), where \( v_j = 1 \) indicates successful verification and \( v_j = -1 \) denotes failure. The resulting sequence \( \{v_{i+1}, \dots, v_{T-1}\} \) provides a fine-grained step-level correctness evaluation for the entire reasoning trace. These step-level signals are used during rollout to compute the aggregated reward \( u \) (see \S~\ref{sec:aggregation}). Unlike LLM-based self-evaluation, which often overestimates fluent but invalid reasoning, this fidelity-aware mechanism grounds supervision in objective, tool-based verification.

While WA is used in our experiments due to its strong mathematical solving capabilities, such as equation solving and equivalence checking, our verification module is tool-agnostic. It supports integration with alternatives like SymPy~\cite{meurer2017sympy} or domain-specific solvers. This modular design ensures that GroundedPRM generalizes across reasoning domains while maintaining high verification precision.

\subsection{Hybrid Reward Aggregation}
\label{sec:aggregation}

To construct reward signals that are both verifiable and forward-looking, GroundedPRM introduces a hybrid aggregation mechanism that combines step-level verification with trajectory-level outcome assessment. This design balances two supervision objectives: (1) factual fidelity of intermediate reasoning steps, and (2) global correctness of the final answer.

Given a simulated reasoning trace of length \( T \), we collect step-level correctness signals \( \{v_{i+1}, \dots, v_{T-1}\} \), where each \( v_i \in \{-1,1\} \) is obtained via external tool verification (see \S~\ref{sec:verification}). In addition, we evaluate the final answer against ground truth to obtain a binary outcome signal \( F \in \{-1,1\} \). These signals are aggregated into a single scalar reward:
\begin{align}
u_i = \frac{1}{T-1-i} \sum_{j=i+1}^{T-1} d_j \cdot v_j + \beta \cdot F,
\end{align}
where \( \beta \geq 0 \) is a weighting coefficient that adjusts the contribution of final answer correctness relative to step-level reliability. The resulting reward \( u \) is used during backpropagation in MCTS (see \S~\ref{sec:mcts}) to update value estimates and guide exploration. We further define the MCTS value estimate at each state–action pair $(s_i,a_i)$ as: $Q(s_i, a_i) = u_i + v_i$.

By fusing local and global correctness signals, this hybrid reward formulation offers more stable and interpretable supervision than prior MC-based methods that rely solely on rollout success. Moreover, this mechanism directly addresses the three core limitations of existing PRMs: it reduces reward noise via structure-aware simulation, avoids unverifiable supervision through external tool-based validation, and aligns the reward objective with both step-wise precision and task-level success.

\subsection{Generative Process Reward Model}
\label{sec:generative}
GroundedPRM adopts a generative reward modeling paradigm, enabling seamless integration with instruction-tuned LLMs and providing supervision for open-ended reasoning workflows. Each training instance is structured as a rationale-enhanced sequence that pairs intermediate reasoning steps with corresponding verification outcomes and justifications.

Formally, each instance includes: (1) the original problem \( P \); (2) the full reasoning trajectory \( \{s_1, \dots, s_T\} \); (3) binary labels indicating the sign of the aggregated reward, combining tool-verified step fidelity and rollout outcome signals; and (4) natural-language explanations derived from external tool feedback, retained after consistency filtering to align with the verified binary labels.

Unlike conventional discriminative reward models that treat reward prediction as a binary classification task, we train GroundedPRM autoregressively to generate both correctness labels and rationales conditioned on the problem and its intermediate reasoning trace. This generative formulation improves interpretability and enables seamless integration into LLM-based reasoning pipelines.

\subsection{Data Construction for GroundedPRM Training}
To train GroundedPRM, we apply the full supervision framework described above to the MATH dataset~\cite{hendrycks2021measuring}, constructing a reward-labeled dataset with tool-based step-level verification and hybrid scoring. For each problem, the policy model generates intermediate reasoning steps, which are verified using external tools (see \S~\ref{sec:verification}). Each step is labeled based on tool-verified correctness, and the full trajectory is scored using the hybrid reward mechanism introduced in \S~\ref{sec:aggregation}. To ensure coverage and diversity, we adopt a multi-round MCTS rollout strategy that explores both optimal and suboptimal paths. Post-processing includes filtering incomplete, inconsistent, or tool-unverifiable traces, and formatting the final data into a rationale-enhanced generative structure (see \S~\ref{sec:generative}). Each instance includes the problem, a full reasoning trace, correctness labels, and explanations. The resulting dataset contains approximately 40K verified samples, covering a broad spectrum of problem types and reasoning strategies with high tool-verified fidelity. Exact generation and verification prompts are given in Appendix~\ref{app:prompt}.

\section{Experiment}
\label{sec:exp}
\subsection{Experimental Setup}

\paragraph{Benchmarks.}
We evaluate GroundedPRM from two perspectives: its ability to accurately identify erroneous steps within multi-step reasoning processes, and its effectiveness in directly enhancing downstream task performance.

\begin{itemize}
  \item \textbf{ProcessBench~\cite{zheng2024processbench}.} This benchmark evaluates the ability of reward models to supervise step-level reasoning in mathematical problems. Each instance includes an LLM-generated solution with the first incorrect step annotated by human experts. Models are evaluated based on their ability to accurately identify the first faulty step or confirm that all steps are valid, following standard PRM evaluation protocols.

  \item \textbf{Reward-Guided Greedy Search.} To further assess the utility of GroundedPRM in guiding multi-step reasoning, we perform inference-time decoding using a reward-guided greedy strategy. At each generation step, we sample $N = 8$ candidate actions from Qwen2.5-7B-Instruct~\cite{qwen2025qwen25technicalreport} using a temperature of 1, and select the candidate with the highest predicted reward assigned by the PRM. This process is repeated iteratively until a complete solution is generated. We evaluate this procedure on six mathematical benchmarks: AMC23~\cite{aimo_validation_amc}, AIME24~\cite{aimo-validation-aime}, MATH~\cite{hendrycks2021measuring}, College MATH~\cite{tang2024mathscale}, OlympiadBench~\cite{he2024olympiadbench}, and Minerva MATH~\cite{lewkowycz2022solving}. We also report the result of pass@n, i.e., the proportion of test samples where any of the n samplings lead to the correct final answers.
\end{itemize}

\paragraph{Baselines.}
For both ProcessBench and reward-guided greedy search experiments, we compare GroundedPRM against the following representative baselines. These baselines span a diverse set of supervision strategies, including models trained with human-labeled rewards, automated annotations, and hybrid approaches, as well as a range of training data scales.

\begin{itemize}
    \item \textbf{Math-Shepherd}~\cite{wang2023math}: Utilizes MC estimation to perform automated step-level annotation with hard labels.
    \item \textbf{RLHFlow-PRM series}~\cite{dong2024rlhf}:  Includes DeepSeek and Mistral variants, both of which use MC estimation for data generation, but adopt the Direct Preference Optimization (DPO) training paradigm.
    \item \textbf{Math-PSA-7B}~\cite{wang2024openr}: Trained on mixed annotated data, namely PRM800K~\cite{lightman2023let}, Math-Shepherd~\cite{wang2023math}, and generated data following~\cite{luo2024improve}.
    \item \textbf{EurusPRM-series}~\cite{sun2025freeprm}: EurusPRM-Stage1 and EurusPRM-Stage2 construct weakly supervised labels from final outcomes using noise-aware heuristics.
	\item \textbf{Qwen2.5-Math-7B series}~\cite{zheng2024processbench, zhang2025lessons}: 
	Qwen2.5-Math-7B-Math-Shepherd and Qwen2.5-Math-7B-PRM800K are trained with Math-Shepherd~\cite{wang2023math} and PRM800K~\cite{lightman2023let} using Qwen2.5-Math-7B-Instruct~\cite{yang2024qwen2}, respectively.	
    \item \textbf{Llemma-PRM800K-7B}~\cite{sun2024easy}: Utilizes MC estimation to perform automated step-level annotation with hard labels.
    \item \textbf{ReasonEval-7B}~\cite{xia2025evaluating}: Prompt-based model for evaluating step validity and redundancy.
\end{itemize}

\paragraph{Implementation Details.}
All reward models are fine-tuned on step-labeled reasoning trajectories using LoRA~\cite{hu2022lora} for parameter-efficient adaptation. We use Qwen2.5-7B-Instruct~\cite{qwen2025qwen25technicalreport} as the base model. Complete training hyperparameters are listed in Appendix~\ref{app:hyper}.

\subsection{Results on ProcessBench}
\begin{table*}[t]
\centering
\caption{
F1 scores on ProcessBench for models trained with auto-labeled data. Models marked with $^*$ share the same base model: Qwen2.5-Math-7B-Instruct. GroundedPRM achieves the highest average F1, surpassing the strongest existing model, Math-Shepherd-PRM-7B, by 26\% relative improvement while using only 10\% of the training data. All baseline results are directly cited from~\cite{zhang2025lessons}. Oly. denotes OlympiadBench. Full results are provided in Appendix~\ref{app:results}.}
\vspace{5pt}
\resizebox{\linewidth}{!}{
\setlength{\tabcolsep}{4pt}
\renewcommand{\arraystretch}{1.1}
\begin{tabular}{lcccccc}
\toprule
\textbf{Model} & \textbf{\#Sample} & \textbf{GSM8K} & \textbf{MATH} & \textbf{Oly.} & \textbf{Omni-MATH} & \textbf{Avg.} \\
\midrule
RLHFlow-DeepSeek-8B             & 253K     & 38.8 & 33.8 & 16.9 & 16.9 & 26.6 \\
RLHFlow-Mistral-8B              & 273K     & 50.4 & 33.4 & 13.8 & 15.8 & 28.4 \\
Qwen2.5-Math-7B-Math-Shepherd$^*$ & 445K     & \textbf{62.5} & 31.6 & 13.7 & 7.7 & 28.9 \\
EurusPRM-Stage1$^*$             & 453K     & 44.3 & 35.6 & 21.7 & 23.1 & 31.2 \\
EurusPRM-Stage2$^*$             & 230K     & 47.3 & 35.7 & 21.2 & 20.9 & 31.3 \\
Math-Shepherd-PRM-7B            & 445K     & 47.9 & 29.5 & 24.8 & 23.8 & 31.5 \\
\textbf{GroundedPRM}                 & \textbf{40K}      & 43.4 & \textbf{47.0} & \textbf{33.8} & \textbf{34.4} & \textbf{39.7} \\
\bottomrule
\end{tabular}}
\label{tab:mix-auto-anno-results}
\end{table*}

\paragraph{GroundedPRM Achieves Strong Supervision Performance with High Data Efficiency.}

As shown in Tab. \ref{tab:mix-auto-anno-results},  GroundedPRM achieves the highest average F1 score among all PRMs trained with automatically labeled data, outperforming the second-best model, Math-Shepherd-PRM-7B, by a relative improvement of 26\% while using only 10\% training samples. GroundedPRM also ranks first on MATH, OlympiadBench, and Omni-MATH, indicating strong capability in evaluating complex mathematical reasoning steps. These results reinforce our central hypothesis: verifiable, structure-guided supervision is substantially more effective than scale alone. GroundedPRM’s fidelity-aware rewards, grounded in tool-based validation and MCTS-based credit assignment, enable efficient learning under low-resource constraints.

\paragraph{Generative Supervision Enhances Interpretability and Robust Generalization.}
Unlike prior PRMs that produce only binary decisions, GroundedPRM adopts a generative format that outputs both a step-level reward and an accompanying rationale. This design improves alignment with instruction-tuned LLMs, encourages interpretable supervision, and enables the model to better distinguish between fluent but incorrect reasoning and truly valid logic. Empirically, GroundedPRM achieves notable improvements on challenging benchmarks like OlympiadBench and MATH, where fine-grained error localization is essential. These results suggest that explanation-based rewards foster more robust and generalizable reasoning behavior.

\subsection{Analysis and Discussions}

\paragraph{GroundedPRM Provides Superior Data Efficiency through Structured and Fidelity-Aware Supervision.}

\begin{table*}[t]
\centering
\caption{
F1 scores of GroundedPRM and Qwen2.5-Math-7B-PRM800K under matched training sizes.
Both methods are trained using Qwen2.5-7B-Instruct but differ in supervision sources. Despite relying solely on automatically labeled data, GroundedPRM consistently outperforms Qwen2.5-Math-7B-PRM800K across all data scales. Oly. denotes OlympiadBench.}
\vspace{5pt}
\resizebox{\linewidth}{!}{
\setlength{\tabcolsep}{5pt}
\renewcommand{\arraystretch}{1.1}
\begin{tabular}{l|c|cccc|c}
\toprule
\textbf{\#Sample} & \textbf{Model} & \textbf{GSM8K} & \textbf{MATH} & \textbf{Oly.} & \textbf{Omni-MATH} & \textbf{Avg.} \\
\midrule
\multirow{2}{*}{10K} 
& Qwen2.5-Math-7B-PRM800K      & 30.3 & 31.6 & 21.9 & 19.8 & 25.9 \\
& GroundedPRM     & \textbf{39.0} & \textbf{41.9} & \textbf{29.4} & \textbf{29.8} & \textbf{35.0} \\
\midrule
\multirow{2}{*}{20K} 
& Qwen2.5-Math-7B-PRM800K      & 37.4 & 32.9 & 29.9 & 30.6 & 32.7 \\
& GroundedPRM     & \textbf{39.9} & \textbf{44.0} & \textbf{30.1 }& \textbf{31.4} & \textbf{36.4} \\
\midrule
\multirow{2}{*}{30K} 
& Qwen2.5-Math-7B-PRM800K      & 37.5 & 40.0 & 28.4 & \textbf{34.8} & 35.2 \\
& GroundedPRM     & \textbf{42.1} & \textbf{47.4} & \textbf{30.7} & 31.7 & \textbf{38.0} \\
\midrule
\multirow{2}{*}{40K} 
& Qwen2.5-Math-7B-PRM800K      & 43.1 & 46.0 & 32.9 & 34.0 & 39.0 \\
& GroundedPRM     & \textbf{43.4} & \textbf{47.0} & \textbf{33.8} & \textbf{34.4} & \textbf{39.7} \\
\bottomrule
\end{tabular}}
\label{tab:prm-scaling}
\end{table*}

To compare the effectiveness of our automatically labeled supervision against human-labeled reward models under identical data budgets, we conduct a controlled comparison with the Qwen2.5-PRM series using the same model architecture, i.e., Qwen2.5-7B-Instruct, and matched training sizes. For each training size, we randomly sample a subset of examples to ensure a fair comparison. This setup isolates the effect of supervision quality by ensuring that both methods are evaluated under the same data scale. As shown in Tab.~\ref{tab:prm-scaling}, GroundedPRM consistently achieves higher F1 scores across all training sizes, despite relying entirely on automatically constructed labels.

\paragraph{Dual-Signal Supervision Enhances Data Fidelity and Credit Attribution.}

\begin{table}[t]
    \centering
 \caption{
F1 scores on ProcessBench under different supervision and inference configurations within the GroundedPRM framework.
\textit{Step-Only} and \textit{Outcome-Only} variants remove one supervision source during training, 
while the \textit{Inference w/o Rationale} variant skips rationale generation and outputs correctness labels directly.
All variants share the same model architecture; the full version combines step-level verification, outcome consistency, and rationale generation.
}  
    \setlength{\tabcolsep}{4pt}
    \renewcommand{\arraystretch}{1.1}
    \begin{tabular}{lccccc}
        \toprule
        \textbf{Method} & \textbf{GSM8K} & \textbf{MATH} & \textbf{OlympiadBench} & \textbf{Omni-MATH} & \textbf{Avg.} \\
        \midrule
        \textbf{Step-Only}        & 40.1 & 42.3 & 28.3 & 29.2 & 35.0 \\
        \textbf{Outcome-Only}     & 1.4  & 3.3  & 1.0  & 1.0  & 1.7  \\
        \textbf{Inference w/o Rationale}  & 34.1 & 34.7 & 22.7 & 23.7 & 28.8 \\
        \textbf{GroundedPRM} & \textbf{43.4} & \textbf{47.0} & \textbf{33.8} & \textbf{34.4} & \textbf{39.7} \\

        \bottomrule
    \end{tabular}
    \label{tab:scoring_f1}
\end{table}

To assess the contribution of our dual-signal supervision, we compare GroundedPRM against two ablations: \textit{Outcome-Only} Supervision, which assigns labels based solely on final-answer correctness from MCTS rollouts, and \textit{Step-Only} Supervision, which uses external tool verification without considering global trajectory outcomes. As shown in Tab.~\ref{tab:scoring_f1}, \textit{Outcome-Only} Supervision severely underperforms due to credit misattribution. Correct steps may be penalized if downstream steps fail, while flawed steps may be rewarded if the final answer happens to be correct. \textit{Step-Only} Supervision achieves higher recall but suffers from precision loss, as external math tools can detect surface-level arithmetic errors but often fail to capture deeper logical flaws, resulting in false positives. A detailed example of this failure mode is provided in Appendix~\ref{app:case ablation}. In contrast, GroundedPRM fuses step-level correctness signals with trajectory-level feedback, enabling accurate credit assignment that is grounded in both local fidelity and global reasoning success. This hybrid design achieves the highest average F1, demonstrating the effectiveness of our supervision framework in producing reliable and structurally aligned reward signals.

\textbf{Rationale Generation Enhances Consistency and Long-Horizon Reasoning.}  
To assess the impact of rationale generation, we compare the full GroundedPRM with an \textit{Inference w/o Rationale} variant that directly predicts correctness labels without generating explanations. As shown in Tab.~\ref{tab:scoring_f1}, removing rationales leads to a consistent drop in F1 across all datasets, with larger gaps on more challenging benchmarks such as MATH and OlympiadBench. Generating intermediate justifications helps maintain step-level consistency, stabilize reward attribution, and localize reasoning errors in complex, long-horizon problems. Qualitative examples in Appendix~\ref{app:case vs LLM} further illustrate how rationale generation improves interpretability and factual grounding without compromising predictive accuracy.

\subsection{Results on Reward-Guided Greedy Search}
\begin{table*}[t]
\centering
\caption{Accuracy of reward-guided greedy search using different PRMs to supervise the Qwen2.5-7B-Instruct policy model. GroundedPRM outperforms all PRMs trained with human, mixed, or automated labels, achieving the highest average accuracy. Oly. denotes OlympiadBench.}
\vspace{5pt}
\resizebox{\linewidth}{!}{
\setlength{\tabcolsep}{4pt}
\renewcommand{\arraystretch}{1.1}
\begin{tabular}{lcccccccc}
\toprule
\textbf{Model} & \textbf{\#Sample} & \textbf{AMC23} & \textbf{AIME24} & \textbf{MATH} & \textbf{College} & \textbf{Oly.} & \textbf{Minerva} & \textbf{Avg.} \\
\midrule
pass@1              & - & 50.0 & 10.0 & 73.4 & 48.5 & 30.0 & 29.8 & 40.3 \\
pass@8(Upper Bound) & - & 82.5 & 20.0 & 90.4 & 61.0 & 48.0 & 49.6 & 58.6 \\
\midrule
\multicolumn{9}{l}{\textbf{Reward-Guided Greedy Search (prm@8)}} \\
\midrule
\multicolumn{9}{l}{\textbf{Trained on Human Annotated Data (PRM800K)}} \\
Qwen2.5-Math-7B-PRM800K & 264K & 60.0 & 10.0 & 75.6 & 36.5 & 23.5 & 29.0 & 39.1 \\
Llemma-PRM800K-7B & ~350K & 42.5 & 6.7 & 72.2 & 47.5 & 27.6 & 29.5 & 37.7 \\
ReasonEval-7B & ~350K & 52.5 & 6.7 & 76.0 & 33.8 & 33.8 & 30.0 & 41.9 \\
\midrule
\multicolumn{9}{l}{\textbf{Trained on a Mix of Human and Automated Annotation Data}} \\
Math-PSA-7B & ~860K & 47.5 & 13.3 & 69.8 & 46.0 & 27.6 & 33.5 & 39.6 \\
\midrule
\multicolumn{9}{l}{\textbf{Trained on Automated Annotation Data}} \\
Math-Shepherd-PRM-7B & 445K & 45.0 & 10.0 & 74.8 & 48.5 & 28.0 & 29.0 & 39.2 \\
RLHFlow-DeepSeek-8B & 253K & 50.0 & 6.7 & 74.2 & 48.0 & 30.9 & 27.5 & 39.5 \\
RLHFlow-Mistral-8B & 273K & 37.5 & 13.3 & 74.8 & 50.5 & 29.8 & 30.0 & 39.3 \\
EurusPRM-Stage1 & 453K & 47.5 & 10.0 & 73.0 & 49.0 & 30.1 & 31.0 & 40.1 \\
EurusPRM-Stage2 & 230K & 45.0 & 13.3 & 73.6 & 51.0 & 31.6 & 32.5 & 41.1 \\
\textbf{GroundedPRM} & 40K & 57.5 & 10.0 & 74.8 & 49.0 & 31.3 & 32.5 & \textbf{42.4} \\
\bottomrule
\end{tabular}}
\label{tab:greedy-results}
\end{table*}

As shown in Tab.~\ref{tab:greedy-results}, GroundedPRM, trained on only 40K automatically labeled examples, achieves the highest average accuracy across all PRMs, surpassing those trained on automated, mixed, or even large-scale human annotations. 
Within the automated annotation group, GroundedPRM achieves new state-of-the-art results on AMC23 and performs on par or better than all counterparts on MATH and Minerva. 
These results validate the effectiveness of the design: tool-grounded verification improves label fidelity, tree-guided path construction yields stable and attribution-aware credit assignment, and rationale-enhanced supervision delivers precise and verifiable step-level evaluation.
By evaluating each candidate step with grounded feedback, GroundedPRM reliably guides the policy toward accurate multi-step reasoning without requiring external demonstrations or value-based lookahead.

\section{Conclusion}
We introduced GroundedPRM, a tree-guided and fidelity-aware framework for process supervision. By combining structured path exploration via MCTS, tool-based step-level verification, hybrid reward aggregation, and rationale-enhanced supervision formatting, GroundedPRM addresses three core limitations of prior PRMs: low factual fidelity, noisy reward signals, and misalignment with step-level reasoning objectives. GroundedPRM is trained on only 40K automatically labeled samples, amounting to just 10\% of the data used by the best-performing PRM trained with auto-labeled supervision. Nevertheless, it achieves up to a 26\% relative improvement in average performance on ProcessBench. When used for reward-guided greedy search, GroundedPRM outperforms even PRMs trained with human-labeled supervision. These results underscore the effectiveness of structured, verifiable reward modeling in enhancing the reasoning capabilities of LLMs.

\section{Future Work}
While GroundedPRM establishes a strong foundation for fidelity-aware and tree-guided process reward modeling, several natural extensions remain. Scaling the underlying LLM may further improve the quality and diversity of generated reasoning paths. Expanding the set of external verifiers beyond the mathematical tool used in this work could enhance flexibility and extend applicability across different reasoning domains. GroundedPRM is inherently tool-agnostic: a “tool” broadly refers to any fidelity verifier that provides execution-grounded feedback for intermediate reasoning steps, including model-based self-checkers, retrieval-augmented verifiers, and rule-based evaluators. Additionally, integrating human preference signals may further align supervision with interpretable and human-consistent reasoning.

A further direction is to integrate GroundedPRM into reinforcement learning pipelines, where it serves as a verifiable reward function guiding policy optimization in long-horizon tasks. Such integration would enable process-level supervision under on-policy updates and reveal how structured rewards interact with exploration, search, and credit assignment. Although our experiments focus on the mathematical domain due to its established PRM benchmarks and baselines, the framework naturally generalizes to any domain where step-level fidelity can be defined and verified, offering a unified and scalable paradigm for grounded process supervision.

\bibliographystyle{plain}
\bibliography{aaai2026}

\newpage
\clearpage
\appendix
\onecolumn
\tableofcontents

\newpage
\section{Algorithm Overview and Pseudocode}
\label{app:code}
To facilitate reproducibility and provide an intuitive understanding of our reward modeling pipeline, we present the high-level pseudocode of GroundedPRM's data generation algorithm. As described in Algo.~\ref{alg:prm}, GroundedPRM integrates MCTS-based reasoning path construction, tool-based step-level verification, and hybrid reward aggregation into a unified supervision framework.

\begin{algorithm}[!htbp]
\caption{Algorithm of GroundedPRM}
\label{alg:prm}
\begin{algorithmic}[1]
\REQUIRE Initial state $s_0$, Node $n$, Node Value $Q$, Execution Round $r$, Max Rounds $R$, Max Children $K$, Tool Result $v$, Rollout Reward $u$, Node List $L \leftarrow \varnothing$, Node Value List $V \leftarrow \varnothing$, Visit Count $\mathcal{N}$

\STATE Initialize: $s_0 \leftarrow s_{\text{initial}}, n \leftarrow n_0, Q_0 \leftarrow 0$

\FOR{$r = 1$ to $R$}
    \WHILE{$n$ is fully expanded}
        \STATE $n_{\text{select}} \leftarrow \text{Selection}(n)$ \COMMENT{Select node with highest UCT score}
        \IF{$n_{\text{select}}$ is terminal or has no children}
            \STATE \textbf{break}
        \ENDIF
        \STATE $n \leftarrow n_{\text{select}}$
    \ENDWHILE

    \STATE $a \leftarrow \text{Generate}(n_{\text{select}})$

    \FOR{$a_j$ from $a_1$ to $a_K$}
        \STATE $s_{i-1} \leftarrow \text{State}(n_{\text{select}})$
        \STATE $n_{i-1} \leftarrow n_{\text{select}}$
        \STATE $s_{i,j} \leftarrow \text{Transition}(s_{i-1}, a_j)$
        \STATE $n_{i,j} \leftarrow \text{Children}(n_{i-1}, s_{i,j})$
        \STATE $Q(s_i) \leftarrow v_{i,j}$ \COMMENT{Verify step with tool}
    \ENDFOR

    \IF{$v_{i,j} = \max(v)$}
        \STATE $n_{\text{sim}} \leftarrow n_{i,j}$
    \ENDIF

    \STATE $n \leftarrow n_{\text{sim}}, s \leftarrow \text{State}(n_{\text{sim}})$

    \WHILE{$n \neq n_{\text{terminal}}$}
        \STATE $s' \leftarrow \text{Transition}(s, a)$
        \STATE $n' \leftarrow \text{Children}(n, s')$
        \STATE $V \leftarrow V + v'$
        \STATE $L \leftarrow L + n'$
        \STATE $n \leftarrow n'$, $s \leftarrow s'$
    \ENDWHILE

    \STATE $F \leftarrow \text{FinalAnswerCorrectness}(n)$ \COMMENT{Compare to ground truth}

    \STATE $T \leftarrow \text{Length}(L)$, $n_i \leftarrow n_{\text{sim}}$

    \FORALL{$v_j \in V$}
        \STATE $u_i \leftarrow \frac{1}{T-1-i} \sum_{j=i+1}^{T-1} d_j \cdot v_j + \beta \cdot F$
    \ENDFOR

    \STATE $Q(s_i, a_i) \leftarrow u_i + v_i$ \COMMENT{Aggregate reward}

    \FOR{$k = i-1$ to $0$}
        \STATE $Q_k \leftarrow Q_k + \gamma^{d_k} \cdot Q(s_i, a_i)$
        \STATE $\mathcal{N}_k \leftarrow \mathcal{N}_k + 1$
    \ENDFOR
\ENDFOR
\end{algorithmic}
\end{algorithm}

\section{Prompt Template for Data Annotation}
\label{app:prompt}
\begin{figure}[!htbp]
    \centering
    \includegraphics[width=0.95\linewidth]{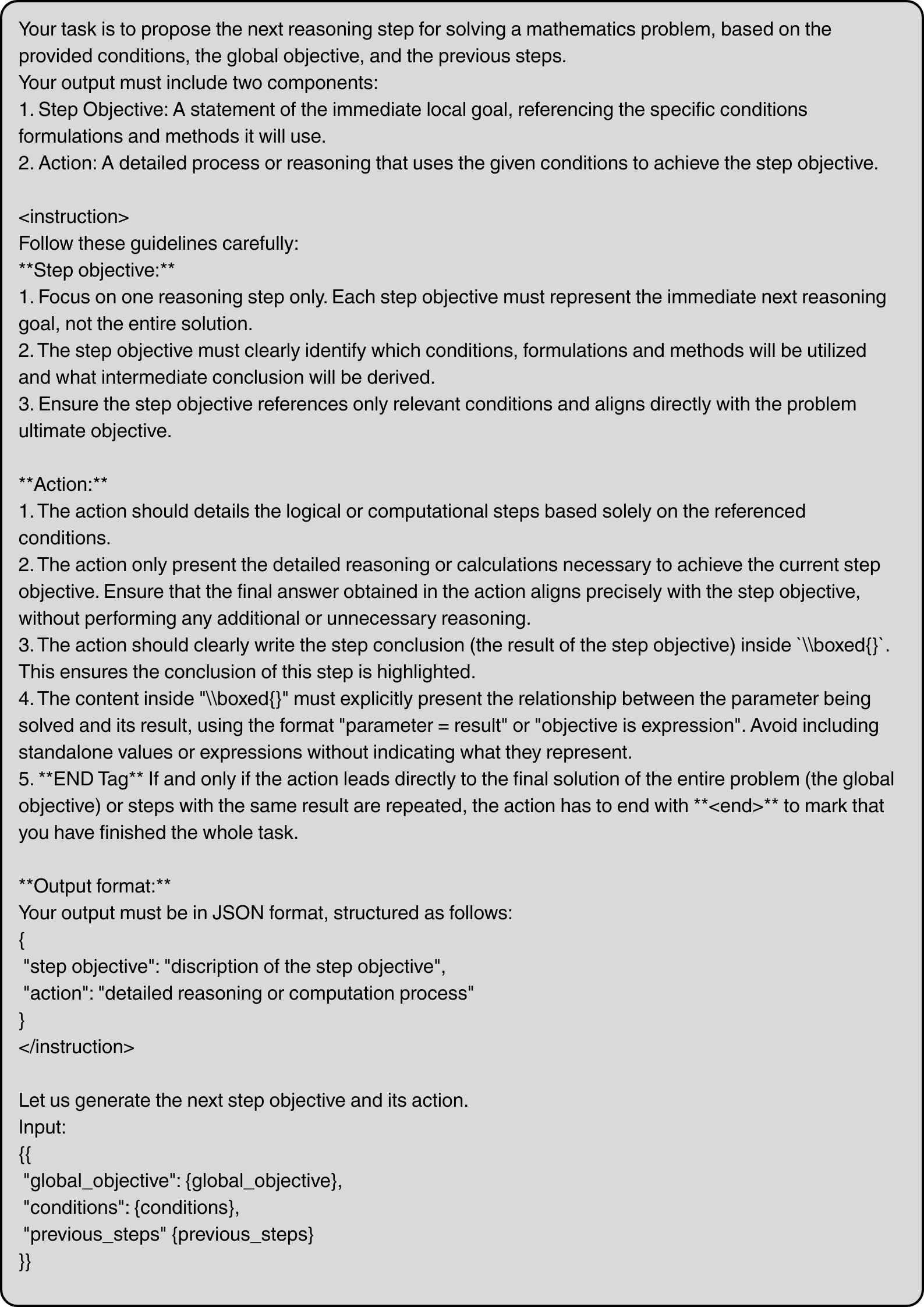}
    \caption{Prompt used to generate the next reasoning step during MCTS rollouts. The output consists of a structured step objective and a logically grounded action aligned with the current goal. These step-level generations are used to construct diverse reasoning trajectories for reward modeling in GroundedPRM.}
    \label{fig:generator}    
\end{figure}

\begin{figure}[!htbp]
    \centering
    \includegraphics[width=0.95\linewidth]{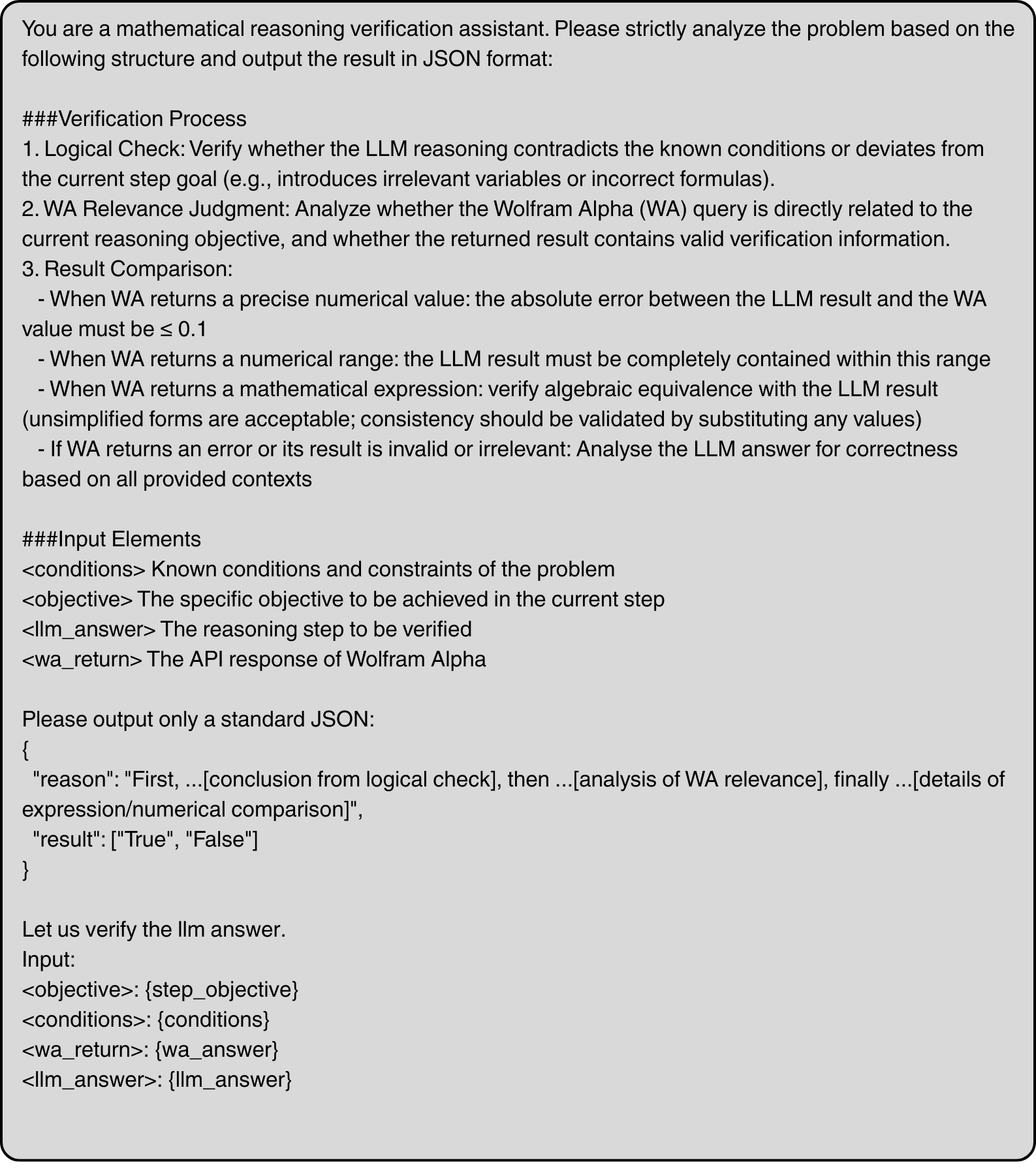}
    \caption{Prompt used for tool-based step-level verification. The assistant analyzes the reasoning step for logical consistency, evaluates the relevance of Wolfram Alpha responses, and outputs a binary correctness label along with a structured rationale, forming the fidelity-aware supervision signal for GroundedPRM.}
    \label{fig:judger}    
\end{figure}

To construct the step-level supervision for GroundedPRM, we adopt two structured prompt templates. The prompt in Fig.~\ref{fig:generator} is used to autoregressively generate intermediate reasoning steps during MCTS rollouts, producing structured trajectories consisting of step objectives and corresponding actions. The prompt in Fig.~\ref{fig:judger} is applied to verify each generated step using external tools, outputting binary correctness labels along with rationale-enhanced explanations, which together form the fidelity-aware supervision signals used to train GroundedPRM.

\section{Training Hyperparameters}
\label{app:hyper}
Tab.~\ref{tab:qwen25_training_config} lists the hyperparameters used to train GroundedPRM. We fine-tuned the model in a sequence-to-sequence manner using the LLaMA-Factory~\citep{zheng2024llamafactory} Trainer implementation on 4×A100 80GB GPUs.

\begin{table}[htbp]
  \centering
     \caption{Training configuration for \texttt{Qwen2.5-7B-Instruct}.}
  \begin{tabular}{@{}ll@{}}
    \toprule
    \textbf{Parameter}                  & \textbf{Value} \\ \midrule
    Model                               & \texttt{Qwen2.5-7B-Instruct} \\
    Torch data type                     & \texttt{bfloat16} \\
    Attention implementation            & flash attention 2 \\
    Lora rank                           & 128 \\
    Lora alpha                         & 256 \\
    Per-device train batch size         & 4 \\
    Gradient accumulation steps         & 8 \\
    Learning rate                       & $3.0\times10^{-5}$ \\
    Number of training epochs           & 6 \\
    LR scheduler type                   & cosine \\
    Max gradient norm                   & 1.0 \\
    Warmup ratio                        & 0.1 \\
    Seed                                & 42 \\
    Optimizer                           & Adam \\
    Gradient checkpointing              & True \\ \bottomrule
  \end{tabular}
  \label{tab:qwen25_training_config}
\end{table}

\section{Supplementary Evaluation Results}
\label{app:results}
We assess the step-level supervision quality of GroundedPRM on ProcessBench, comparing it to several strong PRM baselines trained with automated labels. As shown in Tab. 1 of the main paper, GroundedPRM achieves the highest average F1 score across all benchmarks, with notable gains on MATH, OlympiadBench, and Omni-MATH. These results highlight the effectiveness of our fidelity-aware, structure-guided reward modeling framework in generating accurate and reliable supervision, even under limited data budgets. Full results are provided in Tab.~\ref{tab:full_pb}.

\begin{table*}[!htbp]
\caption{F1 scores on ProcessBench across four math benchmarks. GroundedPRM achieves the highest average F1 score among all PRMs trained with automatically labeled data, outperforming all prior methods by a significant margin, particularly on MATH, OlympiadBench, and Omni-MATH.}
\centering
\small
\resizebox{\linewidth}{!}{
\setlength{\tabcolsep}{5pt}
\renewcommand{\arraystretch}{1.15}
\begin{tabular}{lccc ccc ccc ccc c}
\toprule
\textbf{Scoring Approach} 
& \multicolumn{3}{c}{\textbf{GSM8K}} 
& \multicolumn{3}{c}{\textbf{MATH}} 
& \multicolumn{3}{c}{\textbf{OlympiadBench}} 
& \multicolumn{3}{c}{\textbf{Omni-MATH}} 
& \textbf{Avg. F1} \\
\cmidrule(lr){2-4} \cmidrule(lr){5-7} \cmidrule(lr){8-10} \cmidrule(lr){11-13}
& error & correct & F1 
& error & correct & F1 
& error & correct & F1 
& error & correct & F1 
& \\
\midrule
RLHFlow-PRM-Deepseek-8B       & 24.2 & 98.4 & 38.8 & 21.4 & 80.0 & 33.8 & 10.1 & 51.0 & 16.9 & 10.9 & 51.9 & 16.9 & 26.6 \\
RLHFlow-PRM-Mistral-8B        & 33.8 & 99.0 & 50.4 & 21.7 & 72.2 & 33.4 & 8.2  & 43.1 & 13.8 & 9.6  & 45.2 & 15.8 & 28.4 \\
Qwen2.5-Math-7B-Math-Shepherd & 46.4 & 95.9 & \textbf{62.5} & 18.9 & 96.6 & 31.6 & 7.4 & 93.8 & 13.7 & 4.0 & 95.0 & 7.7 & 28.9 \\
EurusPRM-Stage1               & 46.9 & 42.0 & 44.3 & 33.3 & 38.2 & 35.6 & 23.9 & 19.8 & 21.7 & 21.9 & 24.5 & 23.1 & 31.2 \\
EurusPRM-Stage2               & 51.2 & 44.0 & 47.3 & 36.4 & 35.0 & 35.7 & 25.7 & 18.0 & 21.2 & 23.1 & 19.1 & 20.9 & 31.3 \\
Math-Shepherd-PRM-7B          & 32.4 & 91.7 & 47.9 & 18.0 & 82.0 & 29.5 & 15.0 & 71.1 & 24.8 & 14.2 & 73.0 & 23.8 & 31.5 \\
\textbf{GroundedPRM}         & 31.9 & 67.9 & 43.4 & 36.0 & 67.5 & \textbf{47.0} & 23.4 & 60.5 & \textbf{33.8} & 23.8 & 61.4 & \textbf{34.4} & \textbf{39.7} \\
\bottomrule
\end{tabular}}
\label{tab:full_pb}
\end{table*}

\section{Case Studies}
We conduct qualitative case studies to demonstrate how GroundedPRM performs fine-grained, interpretable supervision across diverse reasoning scenarios. \S~\ref{app:case vs LLM} illustrates its ability to detect arithmetic, algebraic, and constraint-based inconsistencies with high fidelity and structural awareness. \S~\ref{app:case ablation} examines an ablation case that contrasts Step-Only and Dual-Signal supervision, revealing how dual-signal fusion enhances data fidelity and accurate credit attribution.

\subsection{Fidelity- and Error-Type–Aware Reasoning Supervision}
\label{app:case vs LLM}
We present three qualitative cases to illustrate how GroundedPRM delivers fidelity-aware, error-type–aware, and first-wrong-step localization in process supervision. Across all cases, a general-purpose LLM-as-judge fails to catch basic inconsistencies, whereas GroundedPRM recomputes the relevant quantities, checks constraints, and explains why a step is wrong.

\paragraph{Case 1: Basic arithmetic aggregation (Fig.~\ref{case1}).}
A student sums nine quiz scores. The LLM solution totals them to \(\,570\), and the LLM-as-judge accepts this step. In contrast, GroundedPRM reproduces the additions step-by-step \((50 \to 130 \to 210 \to 270 \to 310 \to 400 \to 500 \to 570 \to 630)\), recovers the correct total \(630\), and labels the presented step as incorrect. This shows fidelity-aware arithmetic checking and precise localization of the first wrong step.

\paragraph{Case 2: Sum-of-pairs with spurious halving (Fig.~\ref{case2}).}
The problem gives three pairwise sums of products. The LLM correctly aggregates them to \(210\) but then unjustifiably divides by \(2\) to claim \(105\); the LLM-as-judge still marks the step as correct. GroundedPRM re-evaluates the algebra and confirms \(210\), explicitly naming the first error as a spurious halving and explaining why this single slip corrupts downstream reasoning. This evidences error-type awareness rather than mere outcome comparison.

\paragraph{Case 3: Ratio bounded by inequalities and number-theoretic constraints (Figs.~\ref{case3_1},\ref{case3_2}).}
Given \(b-a=15\) and \(\tfrac{5}{9}<\tfrac{a}{b}<\tfrac{4}{7}\) with \(\gcd(a,b)=1\), the LLM rewrites \(b=a+15\) and proposes candidates; it eventually claims \((a,b)=(26,41)\), which violates the upper bound since \(\tfrac{26}{41}\approx0.6341>\tfrac{4}{7}\approx0.5714\). Fig.~\ref{case3_1} shows the LLM-as-judge validating this wrong candidate by failing to enforce the bound. Fig.~\ref{case3_2} shows GroundedPRM re-checking the inequality numerically, verifying \(b-a=15\) and coprimeness, and recovering a valid pair \((19,34)\) with \(\tfrac{19}{34}\approx0.5588\in(\tfrac{5}{9},\tfrac{4}{7})\). This two-part case highlights constraint checking beyond arithmetic (inequality bounds + number theory).

\paragraph{Takeaways.}
Across the three cases, GroundedPRM (i) recomputes key quantities instead of trusting fluent text, (ii) localizes the first wrong step and names the error type (e.g., spurious halving), and (iii) verifies multi-constraint consistency (inequalities, differences, coprimeness). These behaviors support our central claim: fidelity-aware, structure-guided supervision yields more reliable step-level judgments than generic LLM-as-judge baselines, and produces actionable rationales aligned with process-level reasoning.

\begin{figure}[t]
\centering
\begin{tcolorbox}[width=\textwidth, colback=white, colframe=black!20, boxrule=0.4pt, sharp corners=south, before upper={\footnotesize}]

% ------------------- Question -------------------
\textbf{Question:} 
\begin{tcolorbox}[colback=gray!5, colframe=gray!20, boxrule=0pt, left=4pt, right=4pt, top=3pt, bottom=3pt]

Sally received the following scores on her math quizzes: 50, 80, 80, 60, 40, 90, 100, 70, and 60. Find her mean score.
%\textbf{System prompt:} Your task is to analyse and critique the steps in the solution of a math problem step-by-step.
\end{tcolorbox}
% ------------------- Reasoning Step -------------------

\vspace{6pt}
\textbf{Reasoning Step:}
\begin{tcolorbox}[colback=blue!3, colframe=blue!10, boxrule=0pt, left=6pt, right=6pt, top=3pt, bottom=3pt, before upper={\footnotesize}]
To calculate the mean score of Sally’s math quizzes, first, I need to add up all of her quiz scores.
The question shows the score of each quiz is: 50, 80, 80, 60, 40, 90, 100, 70, and 60.
Adding these together, we get: Total score = 570.
\end{tcolorbox}

% ------------------- GPT-4o (Incorrect) -------------------
\textbf{GPT-4o as Judge:} \textbf{\textcolor{red!60!black}{\faTimesCircle\; **Incorrect judgement.**}}\begin{tcolorbox}[colback=red!3, colframe=red!15, boxrule=0pt, left=6pt, right=6pt, top=3pt, bottom=3pt, before upper={\footnotesize}]
The total sum of the scores is accurate: 50 + 80 + 80 + 60 + 40 + 90 + 100 + 70 + 60 = \textcolor{red!60!black}{570. This step is: \boxed{+}}\\
\end{tcolorbox}

% ------------------- GroundedPRM (Correct) -------------------
\textbf{GroundedPRM as Judge:} \textbf{\textcolor{green!40!black}{\faCheckCircle\; **Correct judgement.**}}
\begin{tcolorbox}[colback=green!3, colframe=green!20, boxrule=0pt, left=6pt, right=6pt, top=3pt, bottom=3pt, before upper={\footnotesize}]
The objective is to find Sally's mean score from her math quizzes. The given scores are 50, 80, 80, 60, 40, 90, 100, 70, and 60. First, I need to verify if the LLM's reasoning correctly adds up all these scores. The LLM states that the total score is 570. Let me add them one by one to check. Starting with 50, adding 80 gives 130. Then, adding another 80 brings it to 210. Adding 60 makes it 270. Next, adding 40 results in 310. Adding 90 increases it to 400. Then, adding 100 brings it to 500. Adding 70 makes it 570, and finally, adding 60 gives 630. \textcolor{green!40!black}{The correct total should be 630, not 570. So the LLM's answer is incorrect here.
According to the conclusion of the judgement, the label is: \boxed{-}}\\
\end{tcolorbox}
\end{tcolorbox}

\caption{Comparison between GPT-4o and GroundedPRM as judge models. GPT-4o \textcolor{red!70!black}{\faTimesCircle\; incorrectly validates} the miscalculated reasoning step (570 instead of 630), while GroundedPRM \textcolor{green!40!black}{\faCheckCircle\; correctly re-evaluates} the arithmetic, detects the error, and provides a rationale-supported supervision label, demonstrating fidelity-aware verification.}
\label{case1}
\end{figure}

\begin{figure}[t]
\centering
\begin{tcolorbox}[width=\textwidth, colback=white, colframe=black!20, boxrule=0.4pt, sharp corners=south, before upper={\footnotesize}]

% ------------------- Question -------------------
\textbf{Question:} 
\begin{tcolorbox}[colback=gray!5, colframe=gray!20, boxrule=0pt, left=4pt, right=4pt, top=3pt, bottom=3pt]
Alice thinks of four positive integers $a \leq b \leq c \leq d$ satisfying $\{a b+c d, a c+b d, a d+b c\}=\{40,70,100\}$. What are all the possible tuples $(a, b, c, d)$ that Alice could be thinking of?
%\textbf{System Prompt:} Analyse and critique the solution steps, give “+” for correct and “–” for incorrect.
\end{tcolorbox}

% ------------------- Reasoning Step -------------------
\textbf{Reasoning Step:}
\begin{tcolorbox}[colback=blue!3, colframe=blue!10, boxrule=0pt, left=6pt, right=6pt, top=3pt, bottom=3pt]
Given the equations \(ab + cd = 40\), \(ac + bd = 70\), and \(ad + bc = 100\), we need to find the positive integers \(a, b, c, d\) such that \(a \leq b \leq c \leq d\). First, let's add all three equations:
\[\ (ab + cd) + (ac + bd) + (ad + bc) = 40 + 70 + 100 \]
\[\ 2(ab + ac + ad + bc + bd + cd) = 210 \]
\[\ ab + ac + ad + bc + bd + cd = 105 \]
\end{tcolorbox}

% ------------------- GPT-4o (Incorrect) -------------------
\textbf{GPT-4o as Judge:} \textbf{\textcolor{red!60!black}{\faTimesCircle\; **Incorrect judgement.**}}
\begin{tcolorbox}[colback=red!3, colframe=red!15, boxrule=0pt, left=6pt, right=6pt, top=3pt, bottom=3pt]
The current step appears to be correct. Let's break it down:\\
1. You are given three equations involving products of four integers \( a, b, c, \) and \( d \): \( ab + cd = 40 \), \( ac + bd = 70 \), \( ad + bc = 100 \)\\
2. To simplify the problem, the step suggests adding all three equations together: $(ab + cd) + (ac + bd) + (ad + bc) = 40 + 70 + 100$\\
3. Simplifying the left-hand side: $ab + cd + ac + bd + ad + bc$. The expression $ ab + ac + ad + bc + bd + cd $ indeed correctly accounts for all pairwise products.\\
4. The right-hand side simplifies to:$40 + 70 + 100 = 210$\\
5. Dividing both sides by 2 to account for each pair being counted twice, you get: $ab + ac + ad + bc + bd + cd = 105$\\
\textcolor{red!60!black}{Conclusively, the current step has correctly simplified the sum of the given equations to derive the result $ ab + ac + ad + bc + bd + cd = 105 $. This expression will be useful in further exploring potential solutions for the tuple $ (a, b, c, d) $. This step is: \boxed{+}\\}
\end{tcolorbox}

% ------------------- GroundedPRM (Correct) -------------------
\vspace{4pt}
\textbf{GroundedPRM as Judge:} \textbf{\textcolor{green!40!black}{\faCheckCircle\; **Correct judgement.**}}
\begin{tcolorbox}[colback=green!3, colframe=green!20, boxrule=0pt, left=6pt, right=6pt, top=3pt, bottom=3pt]
Alright, I'm trying to verify the reasoning step provided. Let me go through it step by step. First, the problem gives us three equations: $ab + cd = 40, ac + bd = 70, ad + bc = 100$. We need to find positive integers \(a, b, c, d\) with \(a \leq b \leq c \leq d\). The reasoning starts by adding all three equations: $(ab + cd) + (ac + bd) + (ad + bc) = 40 + 70 + 100$, Which simplifies to: $ab + cd + ac + bd + ad + bc = 210$. Then, the reasoning factors this sum into: $(ab + ac + ad + bc + bd + cd) = 105$.\\
I need to check if this step is logically sound. Let's see: Adding the three equations: $ ab+cd = 40, ac + bd = 70, ad + bc = 100$. Adding them together gives: $ ab + cd + ac + bd + ad + bc = 40 + 70 + 100 = 210$. So, the sum of all these terms is indeed 210. The reasoning then states that this sum equals \(ab + ac + ad + bc + bd + cd = 105\). Wait, that seems off because 210 divided by 2 is 105, but the sum is 210, not 105. There's a mistake here.\\
So, the correct equation after adding should be: $ab + ac + ad + bc + bd + cd = 210$. Not 105. Therefore, the reasoning incorrectly halved the total sum, leading to an incorrect conclusion.\\
This error affects the rest of the reasoning, as subsequent steps would be based on an incorrect total sum. Hence, the logical check fails because the step introduces an error in the equation.\\
In summary, the step adds the equations correctly but incorrectly divides the total sum by 2, leading to an incorrect value for the sum of the terms. This makes the reasoning flawed.\\
\textcolor{green!40!black}{The logical check reveals an error in the reasoning step. Adding the three equations correctly gives a total sum of 210, but the reasoning incorrectly calculates this as 105 by dividing by 2. This introduces an inconsistency in the problem-solving process. According to the judgment conclusion, the label of this step is: \boxed{-}}
\end{tcolorbox}
\end{tcolorbox}
\caption{Comparison between GPT-4o and GroundedPRM as judge models. After correctly aggregating the three sums to $210$, the LLM performs an unjustified halving to $105$. GPT-4o \textcolor{red!70!black}{\faTimesCircle\; incorrectly validates} this step. GroundedPRM \textcolor{green!40!black}{\faCheckCircle\; correctly re-evaluates} the arithmetic to $210$, verifies each sub-sum, localizes the first wrong step (the halving), and outputs a rationale-supported negative label, showing fidelity-aware, error-type–aware verification.}
\label{case2}
\end{figure}

\begin{figure}[t]
\centering
\begin{tcolorbox}[width=\textwidth, colback=white, colframe=black!20, boxrule=0.4pt, sharp corners=south, before upper={\footnotesize}]

% ------------------- Question -------------------
\textbf{Question:} 
\begin{tcolorbox}[colback=gray!5, colframe=gray!20, boxrule=0pt, left=4pt, right=4pt, top=3pt, bottom=3pt]
The positive integers $a$ and $b$ have no common divisor larger than 1 . If the difference between $b$ and $a$ is 15 and $\frac{5}{9}<\frac{a}{b}<\frac{4}{7}$, what is the value of $\frac{a}{b}$ ?
%\textbf{System prompt:} Analyse and critique the solution steps, give “+” for correct and “–” for incorrect.
\end{tcolorbox}

% ------------------- Reasoning Step -------------------
\textbf{Reasoning Step:}
\begin{tcolorbox}[colback=blue!3, colframe=blue!10, boxrule=0pt, left=6pt, right=6pt, top=3pt, bottom=3pt]
To find the value of \( \frac{a}{b} \) under the given conditions, let's reason through the problem step-by-step: 

First, understanding the conditions: $ \gcd(a,b) = 1$ means that \(a\) and \(b\) are coprime, i.e., they have no common divisors other than 1. $ b - a = 15$ indicates the difference between \(b\) and \(a\). $\frac{5}{9} < \frac{a}{b} < \frac{4}{7}$ gives us bounds on the ratio \( \frac{a}{b} \).

Second, interpreting the inequality:
The inequality $\frac{5}{9} < \frac{a}{b} < \frac{4}{7} $ implies:
$\frac{a}{b}$ is greater than \( \frac{5}{9} \approx 0.5556 \)
$\frac{a}{b}$ is less than \( \frac{4}{7} \approx 0.5714 \)

Third, simplifying the problem: Since \( b - a = 15 \), we can write \( b = a + 15 \). Substituting this into the inequality gives: $\frac{5}{9} < \frac{a}{a+15} < \frac{4}{7} $

Fourth, solving for \(a\) and \(b\):
We want to find values of \(a\) and \(b\) that satisfy the conditions while being integers. Let's test possible values of \(a\) and \(b\) keeping in mind \(b = a + 15\).

Fifth, finding the solution:
We know from the bounds of the inequality that \(a\) must be close to \(b\) (but smaller) since both fractions are close to each other. However, \(a\) and \(b\) also differ by 15, so the fraction should give a value slightly less than but close to \( \frac{4}{7} \). Testing different values for \(a\) and \(b\), while ensuring they are coprime and \(b-a=15\):

- For \(a=1, b=16\), \( \frac{1}{16} \) is too small.

- For \(a=2, b=17\), \( \frac{2}{17} \) is still too small.

- Continuing this process...

- For \(a=15, b=30\), \( \frac{15}{30} = \frac{1}{2} \) is too large.

- For \(a=21, b=36\), \( \frac{21}{36} = \frac{7}{12} \) is too large.

Notice that when \(a=26\) and \(b=41\), we have:

- \( \frac{26}{41} \) which satisfies the inequality \( \frac{5}{9} < \frac{26}{41} < \frac{4}{7} \) and \( \gcd(26,41) = 1 \).

Sixth, conclusion:
Given the constraints and the testing process, the values \(a=26\) and \(b=41\) satisfy all the conditions. Therefore, the value of \( \frac{a}{b} \) is $\frac{26}{41}$.
\end{tcolorbox}

% ------------------- GPT-4o (Incorrect) -------------------
\textbf{GPT-4o as Judge:} \textbf{\textcolor{red!60!black}{\faTimesCircle\; **Incorrect judgement.**}}
\begin{tcolorbox}[colback=red!3, colframe=red!15, boxrule=0pt, left=6pt, right=6pt, top=3pt, bottom=3pt]
The current step contains an error in calculations and reasoning, which should be addressed for correctness. Let's analyze the problem:\\
1. Given Conditions:

\quad - Positive integers \( a \) and \( b \) with \( \gcd(a, b) = 1 \).

\quad - \( b - a = 15 \).

\quad - \( \frac{5}{9} < \frac{a}{b} < \frac{4}{7} \).

2. Express \( b \) in terms of \( a \):

\quad - Since \( b = a + 15 \), substitute to get \( \frac{a}{a+15} \).

3. Refine the inequality:

\quad - Plug \( b = a + 15 \) into \( \frac{5}{9} < \frac{a}{b} < \frac{4}{7} \), giving \( \frac{5}{9} < \frac{a}{a+15} < \frac{4}{7} \).

4. Estimation by calculation:

\quad - Solve \( \frac{a}{a+15} > \frac{5}{9} \) and \( \frac{a}{a+15} < \frac{4}{7} \) separately to identify candidate values for \( a \)

5. Determine suitable integers:

\quad - The process should ensure integer solutions for \( a \) and \( b \) while maintaining coprimeness and verifying bounds.

Issues in the current step:

\quad - The calculation steps provided for \( a = 21, b = 36 \) yielding \( \frac{21}{36} = \frac{7}{12} \) do not match \( b - a = 15 \) since \( b - a \neq 15\).

\quad - The test case \( a = 26, b = 41 \) accurately satisfies the division \( \frac{26}{41} \) within bounds and has \( b - a = 15 \), as well as ensuring \( \gcd(26, 41) = 1 \). This is the correct solution for confirmed calculations.

\textcolor{red!60!black}{The correct logic requires testing every integer value while ensuring all conditions (\(\gcd, b-a=15\)) are met, leading to the right solution. Thus, \( a = 26 \) and \( b = 41 \), is the correct match. This step is: \boxed{+}}
\end{tcolorbox}
\end{tcolorbox}

\caption{Comparison between GPT-4o and GroundedPRM as judge models. Part 1 of 2 (see Fig.~\ref{case3_2} for part 2). Given $b-a=15$ and $ \tfrac{5}{9}<\tfrac{a}{b}<\tfrac{4}{7}$, the LLM rewrites the bound as $ \tfrac{a}{a+15}$ and then proposes candidates. GPT-4o \textcolor{red!70!black}{\faTimesCircle\; incorrectly validates} the step that accepts $(a,b)=(26,41)$ even though $ \tfrac{26}{41}\approx0.6341> \tfrac{4}{7}$, failing the bound check.}
\label{case3_1}
\end{figure}

% ===== Figure 2: GroundedPRM only =====
\begin{figure}[t]
\centering
\begin{tcolorbox}[width=\textwidth, colback=white, colframe=black!20, boxrule=0.4pt, sharp corners=south, before upper={\footnotesize}]

% ------------------- GroundedPRM (Correct) -------------------
\textbf{GroundedPRM as Judge:} \textbf{\textcolor{green!40!black}{\faCheckCircle\; **Correct judgement.**}}
\begin{tcolorbox}[colback=green!3, colframe=green!20, boxrule=0pt, left=6pt, right=6pt, top=3pt, bottom=3pt]
Okay, I need to verify the LLM's reasoning step based on the given problem. Let me go through each part carefully.\\
First, the objective is to find integers a and b such that their greatest common divisor is 1, their difference is 15, and the fraction a/b lies between 5/9 and 4/7. The LLM suggests testing values where a is close to b but smaller, with b = a + 15.\\
Looking at the examples provided:\\
- a=1, b=16: $1/16 \approx 0.0625$, which is way below $5/9$ ($\approx 0.5556$). So this doesn't fit.\\
- a=2, b=17: $2/17 \approx 0.1176$, still too low.\\
- a=15, b=30: $15/30=0.5$, which is exactly $1/2$, but the upper bound is $4/7 \approx 0.5714$. So it's within the range, but let me check if a and b are coprime. gcd(15,30)=15, not 1. So this doesn't work.\\
- a=21, b=36: $21/36=7/12 \approx 0.5833$, which is above $4/7$. So it's outside the upper bound.\\
- Finally, a=26, b=41: $26/41 \approx 0.6341$, which is above $4/7$. Wait, that's over. But the LLM says it's within the range. That seems incorrect because $4/7 \approx 0.5714$, and $26/41 \approx 0.6341$, which is higher than $4/7$. So maybe there's a mistake here.\\
Wait, perhaps I miscalculated $26/41$. Let me compute it again. $26/41$ is indeed approximately $0.6341$, which is greater than $4/7$ ($\approx 0.5714$). So the LLM's conclusion that $26/41$ is within the range is incorrect. Therefore, the reasoning might have an error.\\
But wait, maybe I made a mistake in the comparison. Let me check $26/41$ against $5/9$ and $4/7$ again. $5/9 \approx 0.5556$, and $4/7 \approx 0.5714$. $26/41 \approx 0.6341$, which is above $4/7$. So the fraction is actually outside the upper bound. Therefore, the LLM's answer might be wrong because it claims $26/41$ is within the range when it's not.\\
However, perhaps I should check if there's another pair that fits better. Maybe a=31, b=46: $31/46 \approx 0.6739$, still above $4/7$. a=36, b=51: $36/51=12/17 \approx 0.7059$, which is even higher. Maybe a=25, b=40: $25/40=5/8=0.625$, still above $4/7$. Hmm, perhaps the correct a and b are lower.\\
Wait, let me try a=16, b=31: $16/31 \approx 0.5161$, which is above $5/9$ but below $4/7$. But $b-a=15$, so $31-16=15$. Now, $\gcd(16,31)=1$, since 16 is $2^4$ and 31 is prime. So this could be a valid solution. Let me check: $16/31 \approx 0.5161$, which is between $5/9 \approx 0.5556$? Wait, no, $0.5161$ is less than $5/9$. So that doesn't work.\\
Alternatively, a=19, b=34: $19/34 \approx 0.5588$, which is above $5/9$ but below $4/7$. Check $\gcd(19,34)=1$. So $19/34 \approx 0.5588$, which is between $5/9$ and $4/7$. So this might be the correct pair.\\
So the LLM's answer with a=26 and b=41 is incorrect because $26/41>4/7$. The correct pair should be $a=19$ and $b=34$, which satisfies all conditions.\\
Therefore, the LLM's reasoning has a logical error in verifying the upper bound of the fraction. It incorrectly concluded that $26/41$ is within the required range when it's actually outside. Hence, the result is false.\\
\textcolor{green!40!black}{The LLM's reasoning incorrectly concludes that $ \frac{26}{41} $ is within the specified range. Upon verification, $ \frac{26}{41} \approx 0.6341$, which exceeds the upper bound of $ \frac{4}{7} \approx 0.5714 $. A correct pair satisfying all conditions is $ a=19 $ and $ b=34 $, yielding $ \frac{19}{34}=\frac{1}{2} $, which is within the range and coprime. According to the conclusion of the judgement, the label is: \boxed{{-}}}
\end{tcolorbox}
\end{tcolorbox}

\caption{Comparison between GPT-4o and GroundedPRM as judge models. Part 2 of 2 (complements Fig.~\ref{case3_1}). For the same case, GroundedPRM \textcolor{green!40!black}{\faCheckCircle\; correctly re-evaluates} the inequality, verifies $b-a=15$ and coprimeness, rejects $(26,41)$, and identifies the valid pair $(19,34)$ with $ \tfrac{19}{34}\approx0.5588$; it outputs a rationale-supported negative label, demonstrating fidelity-aware, error-type–aware verification.}
\label{case3_2}
\end{figure}

\subsection{Dual-Signal Supervision Improves Data Fidelity and Credit Attribution}
\label{app:case ablation}

We illustrate the impact of dual-signal supervision through a math factorization example where a Step-Only verifier fails to detect a missing factor. By combining step-level verification with trajectory-level feedback, GroundedPRM identifies the omission and corrects the reasoning, demonstrating its advantage in aligning local fidelity with global consistency.

\paragraph{Case 4: Prime factorization with an omitted factor (Figs.~\ref{case4_1},\ref{case4_2}).}
An LLM computes the GCF of \(180\) and \(168\) but omits the factor \(3\) in \(168\) (using \(2^3\times 7\)) and concludes \(252\); a Step-Only supervisor validates this step. 
In contrast, GroundedPRM re-derives \(168=2^3\times 3\times 7\), checks common primes with \(180=2^2\times 3^2\times 5\), and recovers the correct GCF \(2^2\times 3=12\), labeling the presented step as incorrect. 
This highlights how dual-signal supervision prevents surface-level false positives and yields faithful credit attribution.

\begin{figure}[t]
\centering
\begin{tcolorbox}[width=\textwidth, colback=white, colframe=black!20, boxrule=0.4pt, sharp corners=south, before upper={\footnotesize}]

% ------------------- Question -------------------
\textbf{Question:}
\begin{tcolorbox}[colback=gray!5, colframe=gray!20, boxrule=0pt, left=4pt, right=4pt, top=3pt, bottom=3pt]
 Find the greatest common factor of 180 and 168.
\end{tcolorbox}

\vspace{6pt}
% ------------------- Reasoning Step -------------------
\textbf{Reasoning Step:}
\begin{tcolorbox}[colback=blue!3, colframe=blue!10, boxrule=0pt, left=6pt, right=6pt, top=3pt, bottom=3pt, before upper={\footnotesize}]
To find the greatest common factor (GCF) of 180 and 168, we can use prime factorization. 
First, let's express each number as a product of its prime factors: 
First, \(180 = 2^2 \times 3^2 \times 5\). 
Second, \(168 = 2^3 \times 7\). 
\end{tcolorbox}

% ------------------- Step-Only (Incorrect) -------------------
\textbf{Step-Only:} \textbf{\textcolor{red!60!black}{\faTimesCircle\; **Incorrect judgement.**}}
\begin{tcolorbox}[colback=red!3, colframe=red!15, boxrule=0pt, left=6pt, right=6pt, top=3pt, bottom=3pt, before upper={\footnotesize}]
Okay, so I need to verify the LLM's answer about finding the GCF of 180 and 168 using prime factorization. Let me start by understanding what the problem is asking. The objective is to find the greatest common factor of these two numbers. 

The LLM provided the prime factorizations: 180 is broken down into \(2^2 \times 3^2 \times 5\) and 168 into \(2^3 \times 7\). I should check if these factorizations are correct because that's the foundation for finding the GCF.

Starting with 180: Let's do the prime factorization step by step. 180 divided by 2 is 90, which divided by 2 again is 45. Then 45 divided by 3 is 15, and 15 divided by 3 is 5, which is a prime number. So yes, that gives us \(2^2 \times 3^2 \times 5\), which matches the LLM's answer.

Now for 168: Let's factorize it. 168 divided by 2 is 84, divided by 2 again is 42, and once more by 2 is 21. 21 divided by 3 is 7, which is prime. So that's \(2^3 \times 3 \times 7\). Wait, the LLM wrote \(2^3 \times 7\), but I think they missed the 3. Hmm, that might be an issue. But wait, in their answer, they only mentioned \(2^3 \times 7\), so maybe they considered only the highest powers present in both? Or perhaps they made a mistake.

But wait, the GCF is found by taking the lowest powers of the common prime factors. So even if 168 has a 3, since 180 doesn't have a 7, the common prime factors are just 2 and 3. So the LLM correctly identified the common factors as \(2^2\) and \(3^1\), leading to a GCF of \(2^2 \times 3 = 4 \times 3 = 12\). However, their factorization of 168 didn't include the 3, which might be a mistake, but since they still arrived at the correct GCF, maybe it's acceptable.

Wait, no, actually, the prime factorization must be accurate. If 168 is indeed \(2^3 \times 3 \times 7\), then the GCF should consider the common primes. So 180 has \(2^2\), \(3^2\), and 5. 168 has \(2^3\), 3, and 7. The common primes are 2 and 3. For 2, the lowest power is \(2^2\), and for 3, it's \(3^1\). So the GCF should be \(2^2 \times 3 = 4 \times 3 = 12\). So even though the LLM's factorization of 168 was incomplete, their conclusion was correct because they only took the common factors. So perhaps it's a minor error in the factorization step but the GCF is still right.

Alternatively, maybe the LLM intended to show only the common factors. Let me double-check. If 168 is \(2^3 \times 7\), then the common factors would be \(2^2\), but then the GCF would be 4, which is incorrect because 3 is also a common factor. So the LLM must have included the 3 in their factorization, even if they didn't mention it in the answer. So perhaps there was a typo, but the actual factorization includes the 3, leading to the correct GCF.

In any case, the key point is that the LLM's answer correctly identifies the GCF as 12, even if there was a slight error in the intermediate steps. Therefore, the logical check passes because the reasoning aligns with the known conditions, and the calculation is correct despite a possible minor miscalculation in the factorization of 168. The LLM correctly used prime factorization to find the GCF of 180 and 168. The prime factors of 180 are \(2^2 \times 3^2 \times 5\), and those of 168 are \(2^3 \times 3 \times 7\). The common factors are \(2^2\) and \(3\), leading to a GCF of 12. \textcolor{red!60!black}{Despite a possible error in the factorization of 168, the final result is correct.According to the judgement conclusion, this step is: \(\boxed{+}\)}
\end{tcolorbox}

\end{tcolorbox}

\caption{Comparison between Step-Only and GroundedPRM. 
Step-Only \textcolor{red!70!black}{\faTimesCircle\; incorrectly validates} a step that is arithmetically valid but logically inconsistent, 
while GroundedPRM \textcolor{green!40!black}{\faCheckCircle\; correctly detects} the missing factor in 168’s prime factorization 
through dual-signal verification, ensuring high-fidelity and logically grounded supervision. Part 1 of 2 (see Fig.~\ref{case4_2} for part 2).}
\label{case4_1}
\end{figure}

\begin{figure}[t]
\centering
\begin{tcolorbox}[width=\textwidth, colback=white, colframe=black!20, boxrule=0.4pt, sharp corners=south, before upper={\footnotesize}]

% ------------------- GroundedPRM (Correct) -------------------
\textbf{GroundedPRM} \textbf{\textcolor{green!40!black}{\faCheckCircle\; **Correct judgement.**}}
\begin{tcolorbox}[colback=green!3, colframe=green!20, boxrule=0pt, left=6pt, right=6pt, top=3pt, bottom=3pt, before upper={\footnotesize}]
Alright, I'm trying to verify the LLM's answer regarding the GCF of 180 and 168 using prime factorization. Let me go through the process step by step.

First, the objective is clear: find the GCF of 180 and 168. The LLM provided the prime factorizations as 180 = 2² × 3² × 5 and 168 = 2³ × 7. 

I need to check if these factorizations are correct. Starting with 180: dividing by 2 gives 90, then another 2 gives 45. So that's 2². Then 45 divides by 3 to get 15, then again by 3 to get 5. So that's 3² and 5¹. So 180's prime factors are indeed 2² × 3² × 5.

Now for 168: dividing by 2 gives 84, then 42, then 21. 21 divides by 3 to get 7. So that's 2³ × 3¹ × 7¹. Wait, the LLM only wrote 2³ × 7. That seems missing the 3. Hmm, so the LLM might have missed including the prime factor 3 in 168's factorization.

But wait, maybe I made a mistake. Let me double-check. 168 divided by 2 is 84, divided by 2 again is 42, divided by 2 again is 21. 21 is 3×7. So yes, it's 2³ × 3 × 7. So the LLM's factorization is incomplete because they only included 2³ and 7, omitting the 3.

However, when finding the GCF, we take the minimum exponents of the common prime factors. Both numbers have 2² and 3², but wait, 168 doesn't have 3². It only has 3¹. So actually, the common prime factors are 2 and 3, with exponents 2 and 1 respectively. Therefore, the GCF should be 2² × 3 = 4 × 3 = 12.

But the LLM didn't include the 3 in their factorization of 168, which could lead to an incorrect GCF. However, in this case, since 168 only has one 3, the GCF would still be 3¹, but the LLM's omission might have confused the process.

Wait, but the LLM's answer is just providing the factorizations, not calculating the GCF yet. So perhaps the issue is more about the accuracy of the factorizations before proceeding. Since the LLM missed the 3 in 168's factorization, their information is incomplete, which could affect the subsequent GCF calculation.

\textcolor{green!40!black}{So, the logical check shows that the LLM's reasoning is flawed because it omitted a prime factor in 168. The factorization should include 3, which is crucial for determining the correct GCF. Therefore, the result should be false because the provided factorizations are incorrect.According to the judgement conclusion, this step is: \(\boxed{-}\)}
\end{tcolorbox}
\end{tcolorbox}

\caption{Comparison between Step-Only and GroundedPRM. 
Step-Only \textcolor{red!70!black}{\faTimesCircle\; incorrectly validates} a step that is arithmetically valid but logically inconsistent, 
while GroundedPRM \textcolor{green!40!black}{\faCheckCircle\; correctly detects} the missing factor in 168’s prime factorization 
through dual-signal verification, ensuring high-fidelity and logically grounded supervision. Part 2 of 2 (complements Fig.~\ref{case4_1}).}
\label{case4_2}
\end{figure}

\end{document}